\documentclass[conference,a4paper]{IEEEtran}
\IEEEoverridecommandlockouts                              

\usepackage{amsmath,amsfonts}
\usepackage{algorithmic}
\usepackage{algorithm}
\usepackage{array}
\usepackage[caption=false,font=normalsize,labelfont=sf,textfont=sf]{subfig}
\usepackage{textcomp}
\usepackage{stfloats} 
\usepackage{url}
\usepackage{verbatim}
\usepackage{graphicx}
\usepackage{cite}
\usepackage{subcaption}

\usepackage{amssymb}
\usepackage{orcidlink}
\usepackage{overpic}

\usepackage{graphicx} 
\usepackage{tabularx}
\usepackage{longtable}
\usepackage{booktabs}
\usepackage{lscape}
\usepackage{multirow}

\usepackage{tikz}
\hypersetup{hidelinks}
\title{\LARGE \bf Geometry-Aware Keyframe Selection for Efficient Gaussian Splatting SLAM}

\author{Phuc Nguyen Xuan$^{1}$, Xiem Hoang Van$^{2}$ and Huu-Hung Nguyen$^{3}$
\thanks{$^{1}$Phuc Nguyen Xuan is with the Institute of System Integration, Le Quy Don Technical University, 10000, Ha Noi, Viet Nam
        {\tt\small phucnx.isi@lqdtu.edu.vn}}%
\thanks{$^{2}$Xiem Hoang Van is with University of Engineering and Technology, Vietnam National University, 10000, Ha Noi, Viet Nam
        {\tt\small Xiemhoang@vnu.edu.vn}}%
\thanks{$^{3}$Huu-Hung Nguyen is with the Artificial Intelligence and Data Science Research Center, Le Quy Don Technical University, 10000, Ha Noi, Viet Nam
        {\tt\small hungnh.isi@lqdtu.edu.vn}}%
}

\begin{document}

\maketitle
\thispagestyle{empty}
\pagestyle{empty}

\begin{abstract}

Redundant keyframes can increase storage consumption and optimization overhead in Gaussian Splatting SLAM. Existing insertion strategies often rely on temporal sampling or individual motion-based criteria, which may not explicitly account for the geometric relationship between a candidate observation and the current reference keyframe. This paper presents a reference-relative keyframe selection framework that evaluates candidate observations using geometric consistency, relative camera motion, and rotation-compensated image displacement. Sparse SuperPoint-LightGlue correspondences are geometrically verified using two-view estimation, and the resulting inlier ratio provides a measure of consistency between the candidate and reference observations. The resulting cues are combined into a utility criterion for selecting keyframes with sufficient geometric consistency and viewpoint variation. Experiments on ReplicaMultiagent and AriaMultiagent show that the proposed method reduces storage consumption by up to 71.0\% and end-to-end runtime by up to 42.8\% compared with Coko-SLAM on the evaluated sequences, while achieving competitive rendering quality. On ReplicaMultiagent, the proposed method reduces average tracking ATE by 21.2-57.9\% compared with Coko-SLAM across the evaluated sequences. An ablation comparing motion-only selection with the proposed criterion further evaluates the effect of incorporating multiple selection cues. Overall, the results show that the proposed keyframe selection strategy can reduce computational overhead while retaining competitive tracking and rendering performance in the evaluated settings.

\end{abstract}


\section{INTRODUCTION}
\label{sec:introduction}

Recent advances in 3D Gaussian Splatting (3DGS)~\cite{kerbl20233d} have enabled high-quality and real-time novel-view synthesis through an efficient point-based scene representation. Building upon this representation, numerous Gaussian Splatting SLAM (GS-SLAM) systems have been proposed for online camera tracking and dense scene reconstruction~\cite{tosi2026nerfs}. More recently, collaborative and multi-agent GS-SLAM frameworks have emerged to improve scalability and scene coverage by allowing multiple agents to jointly construct a shared Gaussian map~\cite{yugay2025magic},~\cite{xu2025mac},~\cite{thomas2025grand},~\cite{yu2025hammer}.

Despite these advances, the computational and memory costs of Gaussian optimization remain a major bottleneck~\cite{tosi2026nerfs},~\cite{navaneet2024compgs}. Since optimizing every incoming frame is infeasible, GS-SLAM systems typically maintain a subset of mapping frames (keyframes) to update the Gaussian representation. However, consecutive frames may provide highly overlapping observations, leading to redundant keyframes and increased storage and optimization costs. Prior studies have also shown that a substantial fraction of inserted keyframes can be redundant, indicating that effective frame selection remains an important challenge~\cite{schmuck2019redundancy}.

Existing GS-SLAM systems adopt several strategies for keyframe selection. Interval-based approaches, such as MAGiC-SLAM~\cite{yugay2025magic} and MAC-Ego3D~\cite{xu2025mac}, insert a keyframe every fixed number of frames. While simple and efficient, such strategies do not explicitly assess the geometric or viewpoint variation of incoming observations. Adaptive methods have therefore been proposed based on motion cues, covisibility, overlap, or learned feature representations. For example, MCN-SLAM~\cite{deng2025mcn} triggers keyframe insertion according to optical flow magnitude, MonoGS~\cite{matsuki2024gaussian} and subsequent GS-SLAM systems employ overlap- or covisibility-based criteria, while CoKo-SLAM~\cite{li2026compact} selects keyframes according to DINOv2~\cite{oquab2023dinov2} feature-space distances. These strategies capture different aspects of frame similarity, overlap, or viewpoint change, but generally rely on a specific selection criterion rather than explicitly combining geometric reliability with complementary viewpoint-change cues.

The usefulness of a mapping frame may not be adequately characterized by a single criterion. In particular, a candidate frame should be considered in terms of whether reliable geometric correspondence with the reference keyframe remains available, whether sufficient viewpoint change has occurred, and whether the observation provides additional spatial variation for map optimization. Relying solely on motion or temporal sampling may therefore retain redundant observations or discard frames that provide useful viewpoint diversity.

\begin{figure*}[thpb]
  \centering
     \includegraphics[width=1\textwidth]{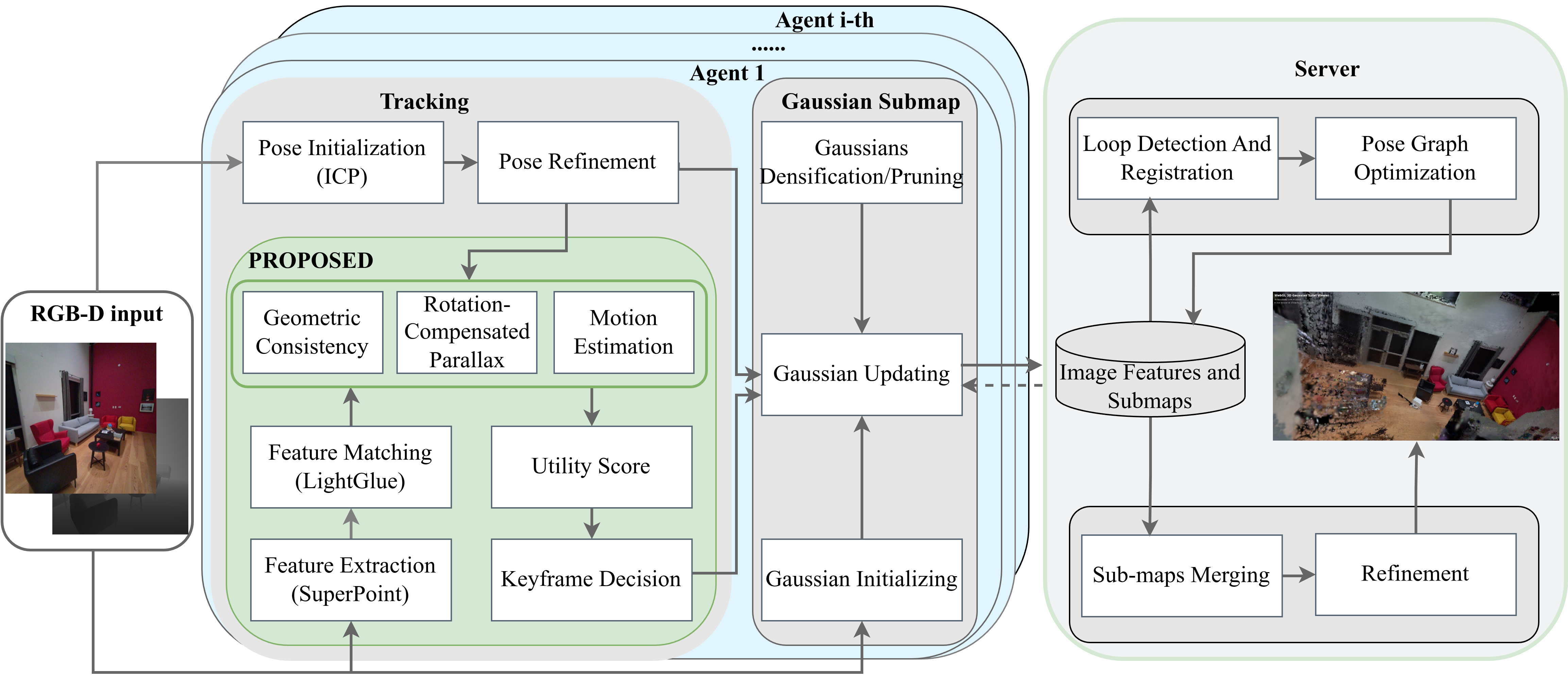}
  \caption{An overview of our approach to keyframing.}
  \label{fig:overview}
\end{figure*}

To address this limitation, we propose a reference-relative keyframe selection strategy for collaborative Gaussian Splatting SLAM (CoGS-SLAM). As illustrated in Fig.~\ref{fig:overview},each candidate frame is evaluated using correspondence-based geometric consistency, relative camera motion, and rotation-compensated image displacement. These complementary measurements are combined into a utility criterion for online keyframe insertion, reducing redundant observations while preserving informative viewpoint changes.

The main contributions are:
\begin{itemize}
\item A reference-relative keyframe selection framework that jointly evaluates geometric consistency and complementary viewpoint-change cues for CoGS-SLAM.
\item A utility criterion combining sparse correspondence verification, relative camera motion, and rotation-compensated image displacement for online keyframe selection.
\item Experiments on ReplicaMultiAgent and AriaMultiAgent demonstrating reduced storage and runtime while maintaining competitive tracking and rendering performance.
\end{itemize}

The remainder of this paper is organized as follows: Section II reviews related work. Section III introduces the proposed method in detail. Section IV presents the experimental evaluation, and Section V concludes the paper.

\section{Related Work}
\label{sec:RelatedWork}
Many recent Gaussian Splatting SLAM (GS-SLAM) systems adopt interval-based keyframing strategies, where a keyframe is inserted every \(n\)-th frame. Examples include MAGiC-SLAM~\cite{yugay2025magic} and MAC-Ego3D~\cite{xu2025mac}. While straightforward to implement, fixed-interval strategies do not explicitly consider the information content of incoming observations and may retain redundant frames, as consecutive views can provide highly overlapping observations~\cite{su2025hpgs}.

To overcome the limitations of fixed keyframe policies, several GS-SLAM systems adopt adaptive selection criteria. MCN-SLAM~\cite{deng2025mcn} triggers keyframe insertion based on average optical flow, which primarily reflects image-space motion but does not explicitly assess geometric correspondence reliability. Other approaches rely on overlap-based measures. MonoGS~\cite{matsuki2024gaussian} estimates Gaussian covisibility between the current frame and the last keyframe and inserts a new keyframe when the overlap falls below a predefined threshold or the camera undergoes sufficient translation, while MGS-SLAM~\cite{zhu2024mgs} follows a similar strategy. SplaTAM~\cite{keetha2024splatam} instead selects keyframes at fixed intervals and uses overlap for keyframe retrieval during map optimization. These approaches capture motion or overlap from different perspectives, but do not explicitly formulate the keyframe decision as a joint utility that combines geometric correspondence reliability with complementary measures of viewpoint change.

More recently, Coko-SLAM~\cite{li2026compact} introduced a feature-space keyframe selection strategy based on pretrained DINOv2~\cite{oquab2023dinov2} embeddings, where a new keyframe is inserted when its feature distance from existing keyframes exceeds a threshold. This provides an appearance-based criterion for identifying visually dissimilar observations, but does not explicitly incorporate geometric correspondence reliability or translation-induced viewpoint variation.

Beyond keyframe insertion, redundancy has also been addressed through map sparsification. Schmuck and Chli~\cite{schmuck2019redundancy} proposed information-theoretic criteria to identify and remove redundant keyframes after map construction, demonstrating that substantial compression can be achieved with limited accuracy loss. In contrast, our work focuses on online redundancy-aware keyframe selection before map insertion, using sparse geometric correspondence to assess reference reliability and complementary motion and parallax cues to characterize viewpoint variation within a unified utility formulation.

\section{Method}
\label{sec:Method}

We formulate online keyframe selection as a reference-relative assessment of observation change. Given a candidate frame $F_c$ and the latest selected keyframe $F_r$, the method does not attempt to predict the subsequent contribution of $F_c$ to Gaussian reconstruction. Instead, it evaluates whether the candidate exhibits sufficient change relative to the current reference to justify insertion.

Three measurements are considered: correspondence-based geometric consistency, relative camera motion, and rotation-compensated image displacement. These measurements provide evidence from correspondence, pose, and image-space domains, respectively, and are combined into a bounded utility score for the insertion decision. Sparse SuperPoint~\cite{detone2018superpoint} and LightGlue~\cite{lindenberger2023lightglue} correspondences are used to obtain the required image-space measurements.

\subsection{Frame Utility Estimation}

\subsubsection{Sparse Feature Correspondence}

For each candidate frame $F_c$, SuperPoint~\cite{detone2018superpoint} extracts image keypoints and descriptors, while LightGlue
~\cite{lindenberger2023lightglue} establishes tentative correspondences with the reference keyframe $F_r$. Only correspondences whose matching confidence exceeds a fixed threshold are retained.

Let
$\{(\mathbf{p}_i^r,\mathbf{p}_i^c)\}_{i=1}^{N_m}$
denote the set of retained tentative correspondences,
where
$\mathbf{p}_i^r=[u_i^r,v_i^r]^\top$
and
$\mathbf{p}_i^c=[u_i^c,v_i^c]^\top$
represent the image coordinates in the reference keyframe and the candidate frames, respectively.

The sparse representation avoids dense image-level processing during frame selection. However, correspondence quality can deteriorate in low-texture, motion-blurred, or highly dynamic regions. Therefore, correspondence failure is treated as an unreliable measurement condition rather than being interpreted directly as evidence of geometric novelty.

\subsubsection{Geometric Consistency Estimation}

The first measurement evaluates whether the current observation remains
geometrically consistent with the reference keyframe. An essential matrix
is estimated from the tentative correspondences using RANSAC. Let
$\mathcal{I}$ denote the set of correspondences classified as geometric
inliers and let

\begin{equation}
N_i = |\mathcal{I}|
\end{equation}

be its cardinality. The geometric consistency score is defined as

\begin{equation}
G_i =
\frac{N_i}{N_m},
\label{eq}
\end{equation}

where $0\leq G_i\leq1$.

A large value of $G_i$ indicates that a large fraction of the matched features is consistent with a common two-view geometry. When the reference and candidate viewpoints remain similar, this provides evidence that the reference keyframe can still explain the current observation and that the candidate is potentially redundant. Conversely, a reduction in $G_i$ indicates weaker geometric consistency with the reference.

Importantly, $G_i$ is not interpreted as a direct measure of geometric novelty. A low inlier ratio may also result from insufficient texture, motion blur, dynamic content, or unreliable feature matching. The geometric term is therefore combined with viewpoint-change measurements rather than being used alone.

\subsubsection{Camera Motion Estimation}

Relative camera motion provides a complementary measurement of viewpoint variation. Given the camera poses of the reference and candidate frames, the relative transformation is computed as
\begin{equation}
\mathbf{T}_{rel}
=
\mathbf{T}_{r}^{-1}
\mathbf{T}_{c},
\end{equation}
where
\begin{equation}
\mathbf{T}_{rel}
=
\begin{bmatrix}
\mathbf{R}_{rel} & \mathbf{t}_{rel}\\
\mathbf{0}^{\top} & 1
\end{bmatrix},
\end{equation}

with $\mathbf{R}_{rel}\in SO(3)$ denoting the relative rotation and $\mathbf{t}_{rel}\in\mathbb{R}^{3}$ the relative translation.

The translation magnitude is $t_i = |\mathbf{t}_{rel}|_2,$ while the rotation magnitude is obtained from the Rodrigues representation:

\begin{equation}
r_i =
\left|
\mathrm{Rodrigues}(\mathbf{R}_{rel})
\right|_2.
\end{equation}

Because translation and rotation have different physical units, they are independently normalized using saturation thresholds:

\begin{equation}
\hat{t}_i =
\min\left(1,\frac{t_i}{\tau_t}\right),
\qquad
\hat{r}_i =
\min\left(1,\frac{r_i}{\tau_r}\right),
\end{equation}

where $\tau_t$ and $\tau_r$ denote the translation and rotation scales used by the keyframe selector. In our implementation, $\tau_t=0.10 \mathrm{m}$,$\tau_r=10^\circ.$

The normalized translation and rotation components are combined as

\begin{equation}
M_i =
\frac{
\sqrt{\hat{t}_i^2+\hat{r}_i^2}
}{
\sqrt{2}
},
\label{eq}
\end{equation}

which gives $M_i\in[0,1]$. A larger $M_i$ indicates a larger change in camera viewpoint according to the adopted translation and rotation scales.

\subsubsection{Rotation-Compensated Viewpoint Variation}

Camera translation can introduce image displacement that is difficult to distinguish from displacement caused by camera rotation. We therefore measure the residual image displacement after compensating for the estimated relative rotation. Following the rotation-compensation principle used in VINS-Mono~\cite{qin2018vins}, a pure-rotation homography is used:
\begin{equation}
\mathbf{H}_{rot}
=
\mathbf{K}
\mathbf{R}_{rel}^{\top}
\mathbf{K}^{-1},
\end{equation}

where $\mathbf{K}$ is the camera intrinsic matrix and $\mathbf{R}_{rel}$ is the relative rotation between the reference and candidate frames.

Each reference feature is warped as:
\begin{equation}
\tilde{\mathbf{p}}_j^r
\sim
\mathbf{H}_{rot}
\mathbf{p}_j^r,
\end{equation}
and the residual displacement between the compensated reference feature and its observed position in the candidate frame is

\begin{equation}
d_i =
\operatorname{median}_{j\in\mathcal{I}}
\left|
\mathbf{p}_j^c-\tilde{\mathbf{p}}_j^r
\right|_2.
\label{eq}
\end{equation}

We use the median rather than the mean to reduce the influence of individual mismatches and localized outliers.

The displacement is normalized by the image diagonal and subsequently bounded using a displacement scale $\tau_p$:

\begin{equation}
P_i =
\min\left(
1,
\frac{d_i}
{\tau_p\sqrt{H^2+W^2}}
\right),
\label{eq}
\end{equation}

where $H$ and $W$ are the image height and width. In our implementation,$\tau_p=0.05.$

The resulting quantity should not be interpreted as physical scene parallax. Instead, it is a rotation-compensated image-displacement proxy
for translation-induced viewpoint variation. This distinction is important because the actual geometric parallax depends on scene depth
and camera translation, whereas Eq.~\eqref{eq} uses only image observations.

\subsection{Redundancy-Aware Keyframe Selection}

The three measurements described above are combined into a bounded utility score:

\begin{equation}
U_i =
w_g(1-G_i)
+
w_m M_i
+
w_p P_i,
\label{eq}
\end{equation}

where $w_g+w_m+w_p=1.$. In our implementation, the weights are fixed to $ w_g=0.5, w_m=0.3, w_p=0.2.$

The asymmetric weighting reflects the different roles of the three measurements. The geometric term receives the largest weight because it characterizes the consistency of the current observation with the reference keyframe. The motion and rotation-compensated displacement terms provide additional evidence of viewpoint variation.

The candidate frame is inserted as a new keyframe when $U_i>\tau$ where the utility threshold is fixed to $\tau=0.45.$

The same parameters are used across the evaluated sequences; no sequence-specific parameter tuning is performed.

The proposed score is therefore an approximation of frame utility rather than a direct measurement of the change in Gaussian reconstruction
quality. Its purpose is to provide an insertion-time decision without requiring additional Gaussian map optimization. 

\subsection{Measurement Validity and Failure Handling}

The utility score is evaluated only when a valid correspondence set and geometric estimate are available. If feature matching produces an insufficient correspondence set or RANSAC fails to obtain a valid geometric model, the utility score is not interpreted as evidence of redundancy. Instead, the candidate frame is conservatively inserted as a keyframe. This policy avoids treating unreliable correspondence as evidence that the current reference adequately represents the candidate observation.

The resulting procedure requires only sparse feature extraction, feature matching, two-view geometric verification, and simple pose-based
computations before a keyframe decision is made. The computational cost of these components and their contribution to the overall system runtime are reported separately in Section~IV.

\section{Experiments}
\label{sec:Experiments}

\subsection{Experimental Setup}

All experiments were conducted on a workstation equipped with an NVIDIA RTX 3060 GPU (12~GB). The proposed method was evaluated on the ReplicaMultiagent~\cite{hu2023cp,straub2019replica} and AriaMultiagent~\cite{pan2023aria} benchmarks, covering Apart-0, Apart-1, Apart-2, and Office-0 from ReplicaMultiagent, and Room-0 and Room-1 from AriaMultiagent.

We compare our approach with two representative baselines: MAGiC-SLAM~\cite{yugay2025magic}, which inserts keyframes at fixed intervals, and Coko-SLAM~\cite{li2026compact}, which employs a learned covisibility-based keyframe selection strategy. Publicly available implementations and recommended parameter settings are used whenever possible.

Performance is evaluated using tracking accuracy (ATE RMSE~\cite{sturm2012benchmark}), rendering quality (PSNR, SSIM~\cite{wang2004image}, LPIPS~\cite{zhang2018unreasonable}, and Depth L1), and computational efficiency (number of selected keyframes, submap size, and wall-clock runtime). Runtime is reported as the maximum execution time across all agents, corresponding to the end-to-end completion time under parallel execution.
All methods use the same mapping and optimization pipeline, with keyframe management as the primary difference.

\subsection{Experimental Results}

We first analyze the efficiency of the proposed strategy in terms of keyframe redundancy, storage consumption, and runtime, followed by evaluations of tracking accuracy and rendering quality.

\subsubsection{Keyframe Efficiency}
Tables~\ref{table:redundance_keyframes_Rep} and~\ref{table:redundance_aria} show that the proposed strategy selects different numbers of keyframes across sequences, with fewer retained frames in most cases. This reduction is accompanied by lower submap storage and shorter end-to-end runtime on both ReplicaMultiagent and AriaMultiagent.

\begin{figure*}[thpb]
  \centering
     \includegraphics[width=0.8\textwidth]{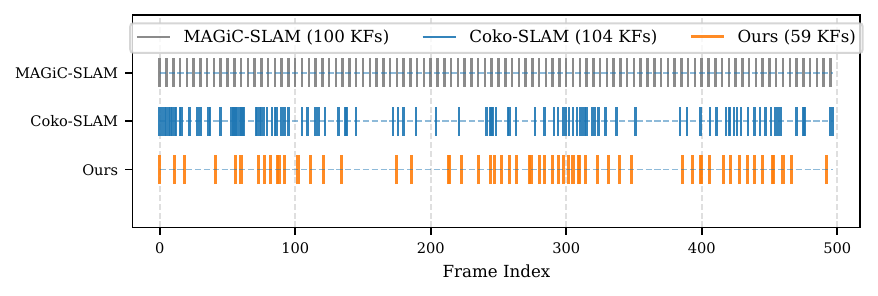}
      \caption{Distribution of selected keyframes on the AriaMultiagent Room-1 Agent-2 sequence. Coko-SLAM inserts keyframes at a relatively dense rate, whereas the proposed utility-driven strategy adaptively retains only informative observations.}
      \label{fig:kf_distribution}
\end{figure*}

As illustrated in Fig.~\ref{fig:kf_distribution}, the selected keyframes are distributed less densely along the trajectory than those of Coko-SLAM~\cite{li2026compact}, reflecting the reference-relative selection criterion. Compared with fixed-interval insertion in MAGiC-SLAM~\cite{yugay2025magic}, the proposed strategy determines keyframe insertion from the measured geometric consistency and viewpoint variation rather than a fixed temporal interval. Despite the reduced keyframe count in most sequences, the resulting tracking and rendering performance remains competitive with the baselines.

\begin{table}[h]
\caption{keyframe efficiency on ReplicaMultiagent~\cite{hu2023cp}}
\label{table:redundance_keyframes_Rep}
\begin{center}
\setlength{\tabcolsep}{3pt}
\begin{tabular}{lclrlrl}
\hline
Dataset & \multicolumn{1}{l}{Methods} & \begin{tabular}[c]{@{}l@{}}KFs\\ (A0/A1)$\downarrow$\end{tabular} & \multicolumn{1}{l}{\begin{tabular}[c]{@{}l@{}}Storage\\(GB) $\downarrow$\end{tabular}} & \begin{tabular}[c]{@{}l@{}}Storage\\ Red.(\%)\end{tabular} & \multicolumn{1}{l}{\begin{tabular}[c]{@{}l@{}}Time\\ (min) $\downarrow$\end{tabular}} & \begin{tabular}[c]{@{}l@{}}Time\\ Red.(\%)\end{tabular} \\ \hline
\multicolumn{1}{c}{Off-0} & Coko-SLAM & 196/235 & 3.30 &  & 178 &  \\
 & Ours & \textbf{60/81} & \textbf{0.96} & \multicolumn{1}{r}{71.00} & \textbf{102} & \multicolumn{1}{r}{42.82} \\ \hline
\multicolumn{1}{c}{Apt-0} & Coko-SLAM & 176/170 & 2.90 &  & 259 &  \\
 & Ours & \textbf{128/142} & \textbf{1.89} & \multicolumn{1}{r}{34.72} & \textbf{184} & \multicolumn{1}{r}{28.82} \\ \hline
\multicolumn{1}{c}{Apt-1} & Coko-SLAM & \textbf{160/148} & 2.70 &  & 258 &  \\
 & Ours & 161/173 & \textbf{2.30} & \multicolumn{1}{r}{14.81} & \textbf{195} & \multicolumn{1}{r}{24.38} \\ \hline
\multicolumn{1}{c}{Apt-2} & Coko-SLAM & 194/175 & 3.10 &  & 259 &  \\
 & Ours & \textbf{117/117} & \textbf{1.61} & \multicolumn{1}{r}{47.94} & 175 & \multicolumn{1}{r}{32.26} \\ \hline
\end{tabular}
\end{center}
\end{table}

\begin{table}[h]
\caption{keyframe efficiency on AriaMultiagent~\cite{pan2023aria}}
\label{table:redundance_aria}
\begin{center}
\begin{tabular}{llcrr}
\hline
\multicolumn{1}{c}{\textbf{Dataset}} & \multicolumn{1}{c}{\textbf{Methods}} & \textbf{KFs(A0/A1/A2)$\downarrow$} & \multicolumn{1}{c}{\textbf{\begin{tabular}[c]{@{}c@{}}Storage\\  (GB) $\downarrow$\end{tabular}}} & \multicolumn{1}{c}{\textbf{\begin{tabular}[c]{@{}c@{}}Time\\ (min) $\downarrow$\end{tabular}}} \\ \hline
\textbf{Room-0} & \textbf{MAGiC-SLAM} & 100/100/100 & 0.72 & 112 \\
 & \textbf{Coko-SLAM} & \textbf{80}/139/125 & 0.67 & 305 \\
 & \textbf{Ours} & 88\textbf{/81/67} & \textbf{0.59} & \textbf{86} \\ \hline
\textbf{Room-1} & \textbf{MAGiC-SLAM} & 100/100/100 & 0.69 & 113 \\
 & \textbf{Coko-SLAM} & 118/86/104 & 0.64 & 271 \\
 & \textbf{Ours} & \textbf{84/76/59} & \textbf{0.58} & \textbf{91} \\ \hline
\end{tabular}
\end{center}
\end{table}

\subsubsection{Tracking Accuracy}

Tables~\ref{table:tracking_replica} and~\ref{table:tracking_aria} report tracking accuracy using ATE RMSE. The proposed keyframe selection achieves comparable or lower tracking error than Coko-SLAM across the evaluated sequences, despite retaining fewer keyframes in most cases. This suggests that the reduced keyframe set does not consistently compromise localization accuracy within the evaluated settings.

\begin{table}[h]
\caption{Tracking results on ReplicaMultiagent~\cite{hu2023cp} (ATE RMSE [cm]$\downarrow$)}
\label{table:tracking_replica}
\begin{center}
\begin{tabular}{llrrrr}
\hline
\multicolumn{1}{c}{\textbf{Methods}} & \multicolumn{1}{c}{\textbf{Agent}} & \multicolumn{1}{c}{\textbf{Apt-0}} & \multicolumn{1}{c}{\textbf{Apt-1}} & \multicolumn{1}{c}{\textbf{Apt-2}} & \multicolumn{1}{c}{\textbf{Off-0}} \\ \hline
\textbf{Coko-SLAM} & \textbf{Agent 0} & 0.19 & 0.57 & 0.62 & 0.51 \\
\textbf{Ours} &  & \textbf{0.18} & \textbf{0.39} & \textbf{0.23} & \textbf{0.23} \\ \hline
\textbf{Coko-SLAM} & \textbf{Agent 1} & 0.46 & 0.45 & 0.52 & 0.46 \\
\textbf{Ours} &  & \textbf{0.33} & \textbf{0.28} & \textbf{0.25} & \textbf{0.21} \\ \hline
\textbf{Coko-SLAM} & \textbf{Average} & 0.33 & 0.51 & 0.57 & 0.49 \\
\textbf{Ours} &  & \textbf{0.26} & \textbf{0.34} & \textbf{0.24} & \textbf{0.22} \\ \hline
\textbf{Impro. (\%)} &  & 21.21 & 33.33 & 57.89 & 55.10 \\ \hline
\end{tabular} 
\end{center}
\end{table}

\begin{table}[h]
\caption{Tracking results on AriaMultiagent~\cite{pan2023aria} (ATE RMSE [cm]$\downarrow$)}
\label{table:tracking_aria}
\begin{center}
\begin{tabular}{llrr}
\hline
\textbf{Methods} & \textbf{Agent} & \multicolumn{1}{l}{\textbf{Room-0}} & \multicolumn{1}{l}{\textbf{Room-1}} \\ \hline
\textbf{Coko-SLAM} & \textbf{Agent 0} & 0.80 & 1.50 \\
\textbf{Ours} & \textbf{} & \textbf{0.68} & \textbf{1.05} \\ \hline
\textbf{Coko-SLAM} & \textbf{Agent 1} & 1.50 & 2.07 \\
\textbf{Ours} & \textbf{} & \textbf{0.95} & \textbf{1.02} \\ \hline
\textbf{Coko-SLAM} & \textbf{Agent 2} & 1.71 & 1.00 \\
\textbf{Ours} & \textbf{} & \textbf{0.71} & \textbf{0.72} \\ \hline
\textbf{Coko-SLAM} & \textbf{Average} & 1.34 & 1.53 \\
\textbf{Ours} & \textbf{} & \textbf{0.78} & \textbf{0.93} \\ \hline
\textbf{Impro. (\%)} &  & \multicolumn{1}{r}{41.79} & \multicolumn{1}{r}{39.22} \\ \hline
\end{tabular}
\end{center}
\end{table}

\subsubsection{Rendering Quality}

Tables~\ref{table:training_view_Replica} and~\ref{table:training_view_Aria} report rendering performance on ReplicaMultiagent and AriaMultiagent, respectively. On ReplicaMultiagent, the proposed method improves or matches the baseline across the reported rendering metrics while retaining fewer keyframes. On AriaMultiagent, the results vary across scenes and metrics, but the proposed method remains competitive despite its lower keyframe count in most evaluated sequences. 

\begin{table}[h]
\caption{Training-view rendering results on ReplicaMultiagent~\cite{hu2023cp}}
\label{table:training_view_Replica}
\begin{center}
\begin{tabular}{llrrrr}
\hline
\textbf{Methods} & \multicolumn{1}{c}{\textbf{Metrics}} & \multicolumn{1}{c}{\textbf{Apt-0}} & \multicolumn{1}{c}{\textbf{Apt-1}} & \multicolumn{1}{c}{\textbf{Apt-2}} & \multicolumn{1}{c}{\textbf{Off-0}} \\ \hline
\multicolumn{1}{c}{\textbf{Coko-SLAM}} & PSNR [dB] $\uparrow$ & 35.25 & 28.63 & 28.12 & 37.87 \\
 & \textbf{LPIPS $\downarrow$} & 0.13 & 0.20 & 0.22 & 0.06 \\
 & \textbf{SSIM  $\uparrow$} & 0.96 & 0.93 & 0.91 & 0.99 \\
 & \textbf{Depth L1[cm]$\downarrow$} & 1.30 & 5.67 & 2.60 & 0.53 \\ \hline
\textbf{Ours} & \textbf{PSNR [dB] $\uparrow$} & \textbf{35.37} & \textbf{29.47} & \textbf{30.63} & \textbf{38.90} \\
 & \textbf{LPIPS $\downarrow$} & \textbf{0.11} & \textbf{0.17} & \textbf{0.17} & \textbf{0.06} \\
 & \textbf{SSIM  $\uparrow$} & \textbf{0.97} & \textbf{0.95} & \textbf{0.96} & \textbf{0.99} \\
 & \textbf{Depth L1[cm]$\downarrow$} & \textbf{0.80} & \textbf{2.53} & \textbf{1.04} & \textbf{0.43} \\ \hline
\end{tabular}
\end{center}
\end{table}

\begin{table}[ht]
\caption{Novel and training view synthesis performance on the AriaMultiAgent dataset.}
\label{table:training_view_Aria}
\centering
\setlength{\tabcolsep}{3pt}
\resizebox{\columnwidth}{!}{%
\begin{tabular}{llrrrcrrr}
\hline
\textbf{Methods} & \textbf{Metrics} 
& \multicolumn{3}{c}{\textbf{Novel Views}} 
& 
& \multicolumn{3}{c}{\textbf{Training Views}} \\
\cline{3-5} \cline{7-9}
& & \textbf{S1} & \textbf{S2} & \textbf{Avg.} 
& 
& \textbf{S1} & \textbf{S2} & \textbf{Avg.} \\
\hline

\textbf{MAGiC-SLAM \cite{yugay2025magic}}
& PSNR [dB] $\uparrow$ 
& 23.45 & \textbf{21.78} & 22.61 
& 
& 24.11 & \textbf{26.17} & 25.14 \\
& SSIM $\uparrow$ 
& 0.89 & \textbf{0.85} & \textbf{0.87} 
& 
& 0.91 & \textbf{0.93} & \textbf{0.92} \\
& LPIPS $\downarrow$ 
& \textbf{0.22} & \textbf{0.21} & \textbf{0.22} 
& 
& \textbf{0.20} & 0.14 & \textbf{0.17} \\
& Depth L1 [cm] $\downarrow$ 
& \textbf{1.33} & 4.96 & \textbf{3.15} 
& 
& \textbf{1.87} & 1.30 & \textbf{1.59} \\
\hline

\textbf{CoKo-SLAM \cite{li2026compact}}
& PSNR [dB] $\uparrow$ 
& 23.06 & 21.72 & 22.39 
& 
& \textbf{24.86} & 25.83 & \textbf{25.34} \\
& SSIM $\uparrow$ 
& 0.84 & 0.82 & 0.83 
& 
& 0.90 & 0.91 & 0.91 \\
& LPIPS $\downarrow$ 
& 0.30 & 0.26 & 0.28 
& 
& 0.25 & 0.21 & 0.23 \\
& Depth L1 [cm] $\downarrow$ 
& 4.24 & 6.09 & 5.16 
& 
& 2.74 & 2.70 & 2.72 \\
\hline

\textbf{Ours}
& PSNR [dB] $\uparrow$ 
& \textbf{24.02} & 21.69 & \textbf{22.86} 
& 
& 24.63 & 25.65 & 25.14 \\
& SSIM $\uparrow$ 
& \textbf{0.89} & 0.84 & 0.86 
& 
& \textbf{0.91} & 0.91 & 0.91 \\
& LPIPS $\downarrow$ 
& 0.24 & 0.23 & 0.23 
& 
& 0.21 & 0.19 & 0.20 \\
& Depth L1 [cm] $\downarrow$ 
& 1.89 & 5.55 & 3.72 
& 
& 2.23 & 2.10 & 2.17 \\
\hline
\end{tabular}%
}

\end{table}

\subsection{Ablation Study}
To evaluate the contribution of the proposed utility formulation, we compare it with a motion-only keyframe selection baseline under comparable resource budgets (Table~\ref{table:ablation}). The motion threshold is varied to obtain a similar level of storage consumption and keyframe retention, allowing the comparison to focus on keyframe quality rather than quantity.

As shown in Table~\ref{table:ablation}, increasing the motion threshold improves localization accuracy in Room-0 but degrades performance in Room-1, indicating that motion-only keyframe selection is highly scene-dependent. In contrast, the proposed utility maintains balanced localization performance across both rooms using a single fixed threshold while operating under a comparable storage budget.

\begin{table}[h]
\caption{Ablation of motion-only and utility-based keyframe selection on AriaMultiagent}
\label{table:ablation}
\begin{center}
\setlength{\tabcolsep}{6pt}
\begin{tabular}{lrlrr}
\hline
\textbf{Method} & \multicolumn{1}{l}{\textbf{$\tau$}} & \textbf{\begin{tabular}[c]{@{}l@{}}Store {[}Gb{]}\\ (R0/R1)\end{tabular}} & \multicolumn{1}{l}{\textbf{\begin{tabular}[c]{@{}l@{}}ATE \\R0{[}cm{]}\end{tabular}}} & \multicolumn{1}{l}{\textbf{\begin{tabular}[c]{@{}l@{}}ATE \\R1{[}cm{]}\end{tabular}}} \\ \hline
Motion-only & 0.60 & 0.62/0.54 & 0.99 & 0.92 \\
Motion-only & 0.65 & 0.63/0.53 & 0.77 & 1.10 \\
Motion-only & 0.70 & 0.53/0.49 & 0.73 & 1.37 \\
Proposed Utility ($U_i$) & 0.45 & 0.59/0.58 & 0.78 & 0.93 \\ \hline
\end{tabular}
\end{center}
\end{table}

\section{Conclusions}
\label{sec:Conclusions}

This paper presented a feature-guided keyframe management framework for CoGS-SLAM. The proposed utility formulation combines correspondence-based geometric consistency, camera motion, and rotation-compensated image displacement to assess reference-relative observation change for online keyframe selection. Experiments on ReplicaMultiAgent and AriaMultiAgent show that the method generally retains fewer mapping frames while reducing submap storage and end-to-end runtime, with competitive tracking and rendering performance across the evaluated sequences. The results further show that motion-only selection can exhibit scene-dependent behavior, while the proposed multi-cue formulation provides a balanced operating point with a fixed configuration.

As the method performs selection at insertion time, it does not account for changes in a keyframe's contribution after subsequent map updates. Future work will therefore consider post-insertion keyframe verification and culling, together with more systematic analysis of cue weighting and threshold selection. Extension to sparse-view and larger-scale collaborative Gaussian Splatting SLAM is also left for future investigation.

\addtolength{\textheight}{-12cm}   









\bibliographystyle{IEEEtran}
\bibliography{refs}  

@article{li2026compact,
  title={Compact Keyframe-Optimized Multi-Agent Gaussian Splatting SLAM},
  author={Li, Monica MQ and Lajoie, Pierre-Yves and Liu, Jialiang and Beltrame, Giovanni},
  journal={arXiv preprint arXiv:2604.00804},
  year={2026}
}

@inproceedings{xu2025mac,
  title={Mac-ego3d: Multi-agent gaussian consensus for real-time collaborative ego-motion and photorealistic 3d reconstruction},
  author={Xu, Xiaohao and Xue, Feng and Zhao, Shibo and Pan, Yike and Scherer, Sebastian and Huang, Xiaonan},
  booktitle={Proceedings of the Computer Vision and Pattern Recognition Conference},
  pages={854--863},
  year={2025}
}

@inproceedings{yugay2025magic,
  title={Magic-slam: Multi-agent gaussian globally consistent slam},
  author={Yugay, Vladimir and Gevers, Theo and Oswald, Martin R},
  booktitle={Proceedings of the Computer Vision and Pattern Recognition Conference},
  pages={6741--6750},
  year={2025}
}

@inproceedings{pan2023aria,
  title={Aria digital twin: A new benchmark dataset for egocentric 3d machine perception},
  author={Pan, Xiaqing and Charron, Nicholas and Yang, Yongqian and Peters, Scott and Whelan, Thomas and Kong, Chen and Parkhi, Omkar and Newcombe, Richard and Ren, Yuheng Carl},
  booktitle={Proceedings of the IEEE/CVF International Conference on Computer Vision},
  pages={20133--20143},
  year={2023}
}

@article{hu2023cp,
  title={Cp-slam: Collaborative neural point-based slam system},
  author={Hu, Jiarui and Mao, Mao and Bao, Hujun and Zhang, Guofeng and Cui, Zhaopeng},
  journal={Advances in Neural Information Processing Systems},
  volume={36},
  pages={39429--39442},
  year={2023}
}

@article{straub2019replica,
  title={The replica dataset: A digital replica of indoor spaces},
  author={Straub, Julian and Whelan, Thomas and Ma, Lingni and Chen, Yufan and Wijmans, Erik and Green, Simon and Engel, Jakob J and Mur-Artal, Raul and Ren, Carl and Verma, Shobhit and others},
  journal={arXiv preprint arXiv:1906.05797},
  year={2019}
}

@inproceedings{sturm2012benchmark,
  title={A benchmark for the evaluation of RGB-D SLAM systems},
  author={Sturm, J{\"u}rgen and Engelhard, Nikolas and Endres, Felix and Burgard, Wolfram and Cremers, Daniel},
  booktitle={2012 IEEE/RSJ international conference on intelligent robots and systems},
  pages={573--580},
  year={2012},
  organization={IEEE}
}

@article{wang2004image,
  title={Image quality assessment: from error visibility to structural similarity},
  author={Wang, Zhou and Bovik, Alan C and Sheikh, Hamid R and Simoncelli, Eero P},
  journal={IEEE transactions on image processing},
  volume={13},
  number={4},
  pages={600--612},
  year={2004},
  publisher={IEEE}
}

@inproceedings{zhang2018unreasonable,
  title={The unreasonable effectiveness of deep features as a perceptual metric},
  author={Zhang, Richard and Isola, Phillip and Efros, Alexei A and Shechtman, Eli and Wang, Oliver},
  booktitle={Proceedings of the IEEE conference on computer vision and pattern recognition},
  pages={586--595},
  year={2018}
}

@inproceedings{schmuck2019redundancy,
  title={On the redundancy detection in keyframe-based slam},
  author={Schmuck, Patrik and Chli, Margarita},
  booktitle={2019 International Conference on 3D Vision (3DV)},
  pages={594--603},
  year={2019},
  organization={IEEE}
}

@article{deng2025mcn,
  title={Mcn-slam: Multi-agent collaborative neural slam with hybrid implicit neural scene representation},
  author={Deng, Tianchen and Shen, Guole and Chen, Xun and Yuan, Shenghai and Shen, Hongming and Peng, Guohao and Wu, Zhenyu and Wang, Jingchuan and Xie, Lihua and Wang, Danwei and others},
  journal={arXiv preprint arXiv:2506.18678},
  year={2025}
}

@article{su2025hpgs,
  title={HPGS-SLAM: Hybrid Point-Guided Dense Visual SLAM With Online Mapping via Gaussian Splatting},
  author={Su, Xin and Zhang, Xiaoang and Pries, Rastin and Steinbach, Eckehard},
  journal={IEEE Robotics and Automation Letters},
  volume={11},
  number={2},
  pages={1882--1889},
  year={2025},
  publisher={IEEE}
}

@inproceedings{matsuki2024gaussian,
  title={Gaussian splatting slam},
  author={Matsuki, Hidenobu and Murai, Riku and Kelly, Paul HJ and Davison, Andrew J},
  booktitle={Proceedings of the IEEE/CVF conference on computer vision and pattern recognition},
  pages={18039--18048},
  year={2024}
}

@article{zhu2024mgs,
  title={Mgs-slam: Monocular sparse tracking and gaussian mapping with depth smooth regularization},
  author={Zhu, Pengcheng and Zhuang, Yaoming and Chen, Baoquan and Li, Li and Wu, Chengdong and Liu, Zhanlin},
  journal={IEEE Robotics and Automation Letters},
  volume={9},
  number={11},
  pages={9486--9493},
  year={2024},
  publisher={IEEE}
}

@inproceedings{keetha2024splatam,
  title={Splatam: Splat track \& map 3d gaussians for dense rgb-d slam},
  author={Keetha, Nikhil and Karhade, Jay and Jatavallabhula, Krishna Murthy and Yang, Gengshan and Scherer, Sebastian and Ramanan, Deva and Luiten, Jonathon},
  booktitle={Proceedings of the IEEE/CVF conference on computer vision and pattern recognition},
  pages={21357--21366},
  year={2024}
}

@article{kerbl20233d,
  title={3d gaussian splatting for real-time radiance field rendering.},
  author={Kerbl, Bernhard and Kopanas, Georgios and Leimk{\"u}hler, Thomas and Drettakis, George and others},
  journal={ACM Trans. Graph.},
  volume={42},
  number={4},
  pages={139--1},
  year={2023}
}

@inproceedings{detone2018superpoint,
  title={Superpoint: Self-supervised interest point detection and description},
  author={DeTone, Daniel and Malisiewicz, Tomasz and Rabinovich, Andrew},
  booktitle={Proceedings of the IEEE conference on computer vision and pattern recognition workshops},
  pages={224--236},
  year={2018}
}

@inproceedings{lindenberger2023lightglue,
  title={Lightglue: Local feature matching at light speed},
  author={Lindenberger, Philipp and Sarlin, Paul-Edouard and Pollefeys, Marc},
  booktitle={Proceedings of the IEEE/CVF international conference on computer vision},
  pages={17627--17638},
  year={2023}
}

@article{oquab2023dinov2,
  title={Dinov2: Learning robust visual features without supervision},
  author={Oquab, Maxime and Darcet, Timoth{\'e}e and Moutakanni, Th{\'e}o and Vo, Huy and Szafraniec, Marc and Khalidov, Vasil and Fernandez, Pierre and Haziza, Daniel and Massa, Francisco and El-Nouby, Alaaeldin and others},
  journal={arXiv preprint arXiv:2304.07193},
  year={2023}
}

@article{thomas2025grand,
  title={GRAND-SLAM: Local Optimization for Globally Consistent Large-Scale Multi-Agent Gaussian SLAM},
  author={Thomas, Annika and Sonawalla, Aneesa and Rose, Alex and How, Jonathan P},
  journal={IEEE Robotics and Automation Letters},
  year={2025},
  publisher={IEEE}
}

@article{yu2025hammer,
  title={Hammer: heterogeneous, multi-robot semantic gaussian splatting},
  author={Yu, Javier and Chen, Timothy and Schwager, Mac},
  journal={IEEE Robotics and Automation Letters},
  year={2025},
  publisher={IEEE}
}

@inproceedings{navaneet2024compgs,
  title={Compgs: Smaller and faster gaussian splatting with vector quantization},
  author={Navaneet, KL and Pourahmadi Meibodi, Kossar and Abbasi Koohpayegani, Soroush and Pirsiavash, Hamed},
  booktitle={European Conference on Computer Vision},
  pages={330--349},
  year={2024},
  organization={Springer}
}

@article{tosi2026nerfs,
  title={How nerfs and 3d gaussian splatting are reshaping slam: a survey},
  author={Tosi, Fabio and Zhang, Youmin and Gong, Ziren and Mattoccia, Stefano and Oswald, Martin R and Sandstrom, Erik and Poggi, Matteo},
  journal={IEEE Transactions on Robotics},
  year={2026},
  publisher={IEEE}
}

@article{qin2018vins,
  title={Vins-mono: A robust and versatile monocular visual-inertial state estimator},
  author={Qin, Tong and Li, Peiliang and Shen, Shaojie},
  journal={IEEE transactions on robotics},
  volume={34},
  number={4},
  pages={1004--1020},
  year={2018},
  publisher={IEEE}
}

\end{document}